\documentclass[letterpaper, 10 pt, conference]{ieeeconf}  

\IEEEoverridecommandlockouts                              

\usepackage{graphicx} 
\usepackage{dblfloatfix}
\usepackage{nidanfloat}
\usepackage{amsmath} 

\title{\LARGE \bf
Adaptive Motion Generation Using \\ Uncertainty-Driven Foresight Prediction
}

\author{Hyogo Hiruma$^{1}$, Hiroshi Ito$^{1}$, and Tetsuya Ogata$^{1}$
\thanks{*This work was supported by JST [Moonshot R\&D][Grant Number JPMJMS2031].}
\thanks{$^{1}$Hyogo Hiruma, Hiroshi Ito and Tetsuya Ogata are with the Department of
              Intermedia Art and Science, Waseda University, Tokyo, Japan
        hiruma@idr.ias.sci.waseda.ac.jp, hiroshi.ito.ws@hitachi.com, ogata@waseda.jp}
}

\begin{document}

\maketitle
\thispagestyle{empty}
\pagestyle{empty}

\begin{figure*}[!b]
    \centering
    \includegraphics[width=\linewidth]{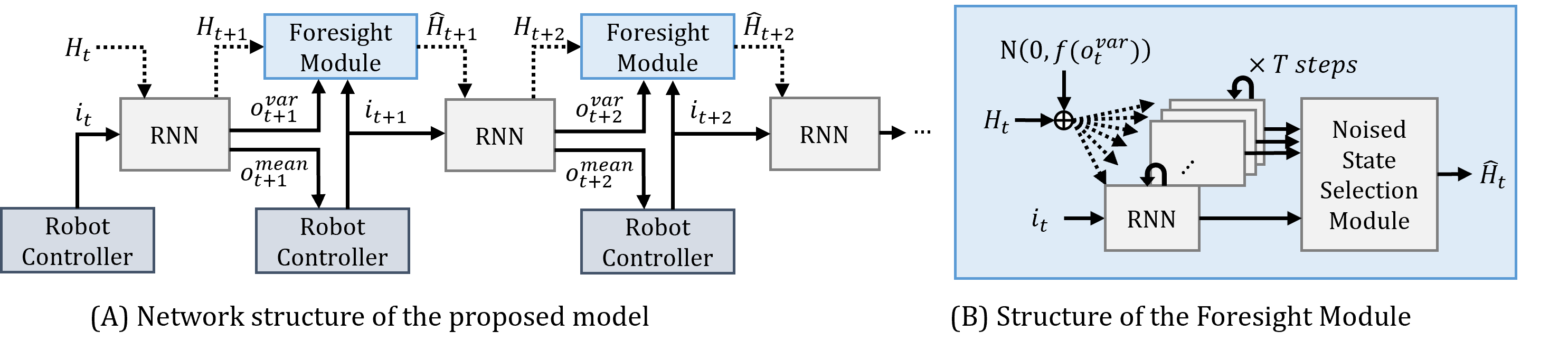}
    \caption{Network structure of the proposed model. (A) The model extends conventional RNN
    models with a foresight module that refines the RNN hidden states through closed-loop predictions.
    (B) The hidden states are refined by selecting the best noise that led to a foresight with the lowest
    expected variance.}
    \label{fig:model_structure}
\end{figure*}

\section{INTRODUCTION}

In the field of robot controlling, the introduction of deep learning methods has enlarged the
application area of robot task execution. One representative approach is learning from
demonstration (LfD), where neural network models are trained to capture the dynamics of robot
tasks through imitation of demonstration data \cite{itosan}. LfD benefits from its
training-efficiency, as robust and flexible robot motions can be acquired from few sets of
demonstrations.

However, task environments may embrace hidden properties that cannot be fully covered
by the demonstrations. One example is a door opening task, where the robot is trained to
open the door without knowing which direction the door could be opened (i.e., pushing
or pulling). As it is difficult for LfD methods to incorporate failure demonstrations,
such as pushing a pull door by mistake, the method's generalization ability becomes limited
in uncertain environments.

Therefore, \textit{adaptiveness} of the robot control model becomes important. This is
because adaptive behaviors are necessary for reducing the uncertainty, which includes
active interaction and feedback collection that aid online policy refinement. Therefore,
the robot control models must be equipped with motivations to (1) correctly understand the
dynamic uncertainty of the environment and to (2) exploratively derive the optimal action. 

Previous research \cite{srnn} proposed a model structure that captures time-dependent
uncertainty of the task, but the deterministic nature of Recurrent Neural Networks (RNN)
limited the derivation of explorative behaviors. Active Inference framework is considered a
promising approach, but existing research required pre-defined uncertainty models and small
actions space, hence suffer in dynamic real-robot applications \cite{ai1}. Although recent
research \cite{pv_rnn, diffusion_policy} succeeded in applying the idea to robot tasks,
their slow inference speed limited the application area.

This paper proposes a motion generation model, based on deep predictive learning framework
\cite{predictive_learning}. The model is designed to be capable of training on tasks that
involve uncertainty, by adding a structure that enhance the applicability against adaptiveness.
Concretely, the proposed model introduces a foresight prediction module to conventional 
RNN models, enables the model to capture accurate uncertainties, which induces adaptive
behaviors at necessary situations.
\section{METHOD}

The proposed model follows the prediction scheme of deep predictive learning. As shown in 
Fig.~\ref{fig:model_structure}(A) the input $i_t$ is the sensor data at the current time step $t$,
and the output $o_{t+1}$ is the expected sensor data at the next time step. The motion is
generated by continuously predicting and applying the next sensor values to the robot controller.

The temporal relationships of sensor data are modeled using a variant of RNNs which incorporate
stochasticity \cite{srnn}. This model assumes that the output can be modeled as a Gaussian distribution,
hence predicts the mean $o_t^{mean}$ and variance $o_t^{var}$ of the expected output. The predicted
variance is considered to reflect the uncertainty of the model, which dynamically change depending
on the task situation.

Fig.~\ref{fig:model_structure}(B) shows the Foresight Module, which is used to refine the hidden states
of RNNs $H_t$ to $\widehat{H}_t$, through closed-loop prediction, namely the foresight prediction. Here,
closed-loop prediction is a method that
performs an internal simulation, by recursively using the self-produced outputs as the next input data.
By continuously predicting in closed-loop manner, the model can create future predictions of multiple
time steps ahead, based on the belief and the dynamics that the model self-structured; the foresights
can be predicted both during and after the training phase. 

The foresights are predicted as follows. First, $H_t$ is applied with $n$ Gaussian noises that are
sampled from $\mathcal{N}(0,\,f\left(o_t^{var}\right))$, where $f\left(\cdot\right)$ normalizes the input
values to $\left[0.05, 0.15\right]$. Then, the noised hidden states $H_t^n$ are used to predict foresights
in $T$ time steps ahead, which produces $n$ simulated predictions $o_{t+T}^{n, mean}$ and $o_{t+T}^{n, var}$.
Finally, $\widehat{H}_t$ is determined by selecting the hidden state that led to a final output with the
lowest expected variance.

\begin{equation}
    \widehat{H}_t = \underset{H_t^n} {\operatorname{argmin}} \; o_{t+T}^{var}
\end{equation}

The proposed model is trained through LfD as in \cite{srnn}, which optimizes the model to maximize
the likelihood of the predicted motions.
\section{EXPERIMENT AND RESULTS}

\begin{figure*}[tbp]
  \begin{minipage}[b]{0.25\linewidth}
    \centering
    \includegraphics[width=\columnwidth]{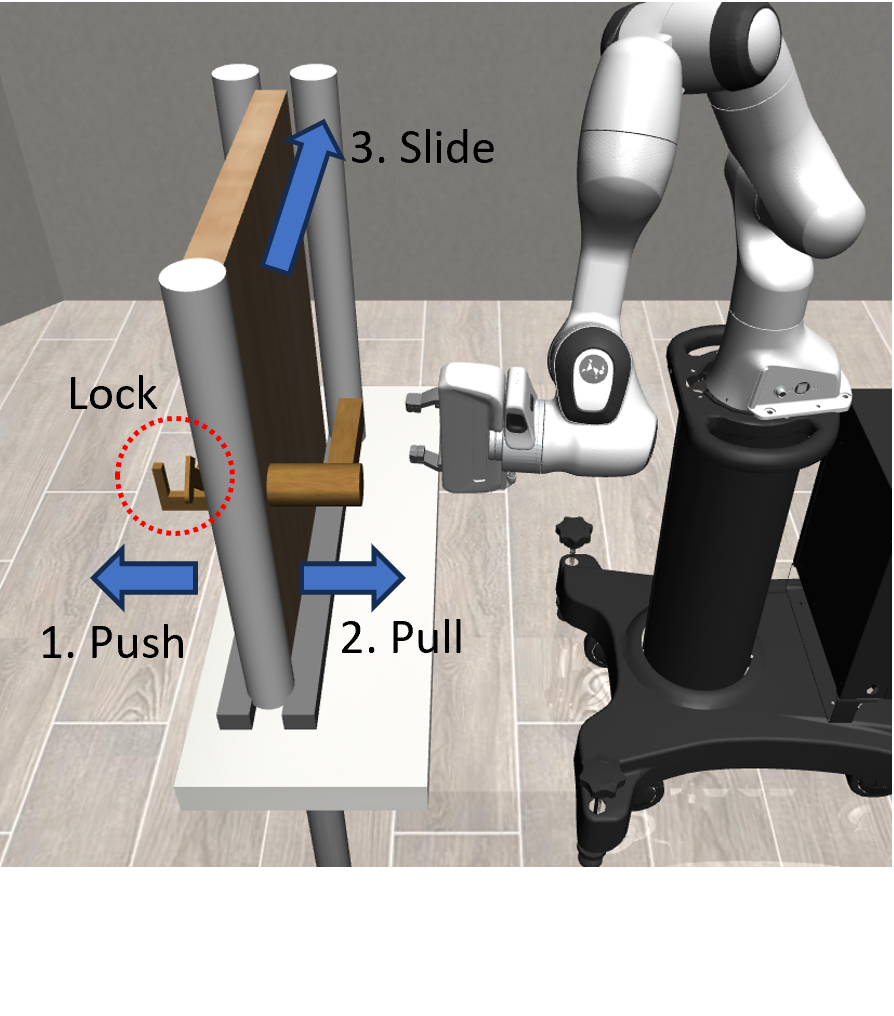}
    \caption{Door opening task environment. The door can
    be opened either by pushing, pulling, or sliding.}
    \label{fig:env}
  \end{minipage}
  \hspace{0.01\columnwidth}
  \begin{minipage}[b]{0.30\linewidth}
    \centering
    \includegraphics[width=\columnwidth]{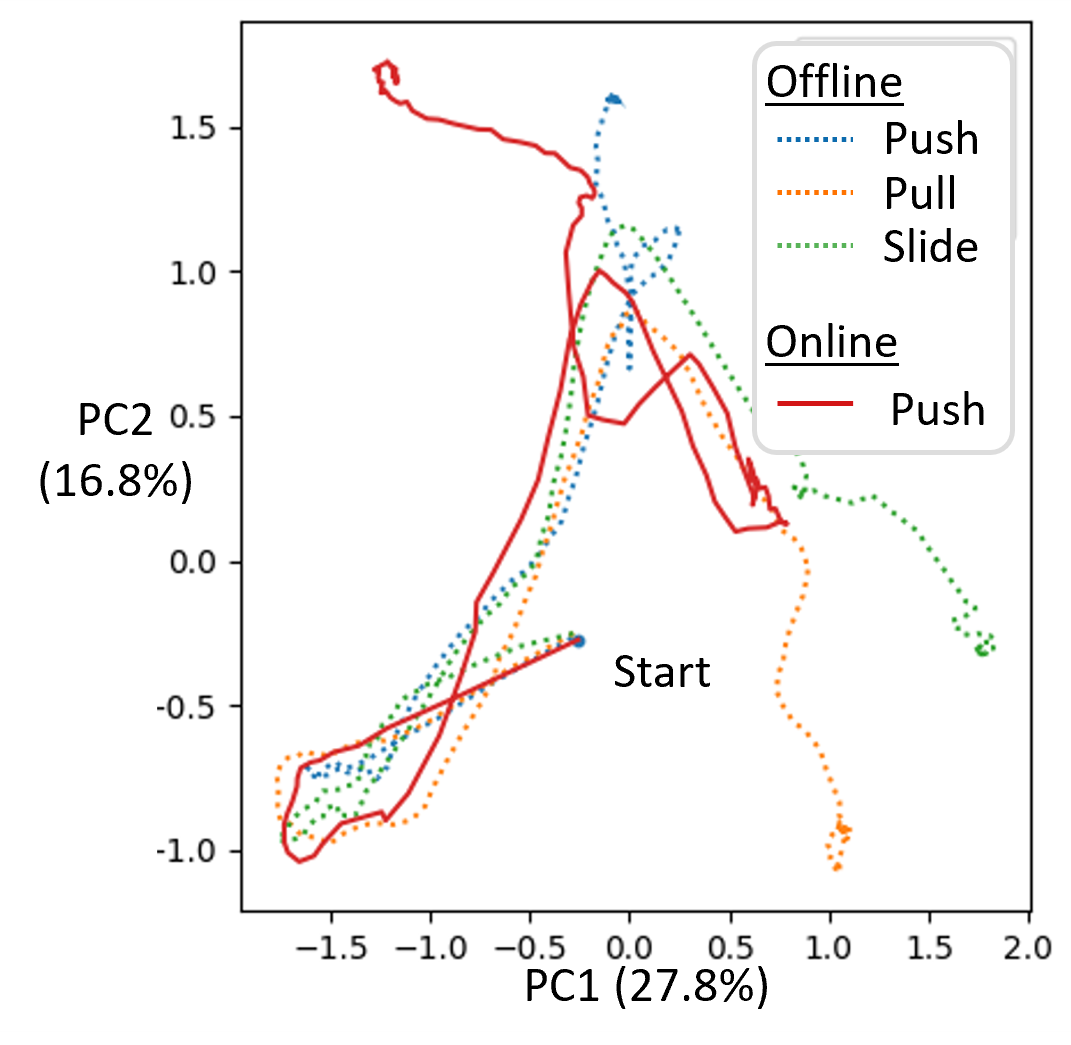}
    \caption{Transitioning RNN hidden states per motion type.
    States are compressed to two dimensions, using
    principal component analysis. }
    \label{fig:pca}
  \end{minipage}
  \hspace{0.01\columnwidth}
  \begin{minipage}[b]{0.40\linewidth}
    \centering
    \includegraphics[width=\columnwidth]{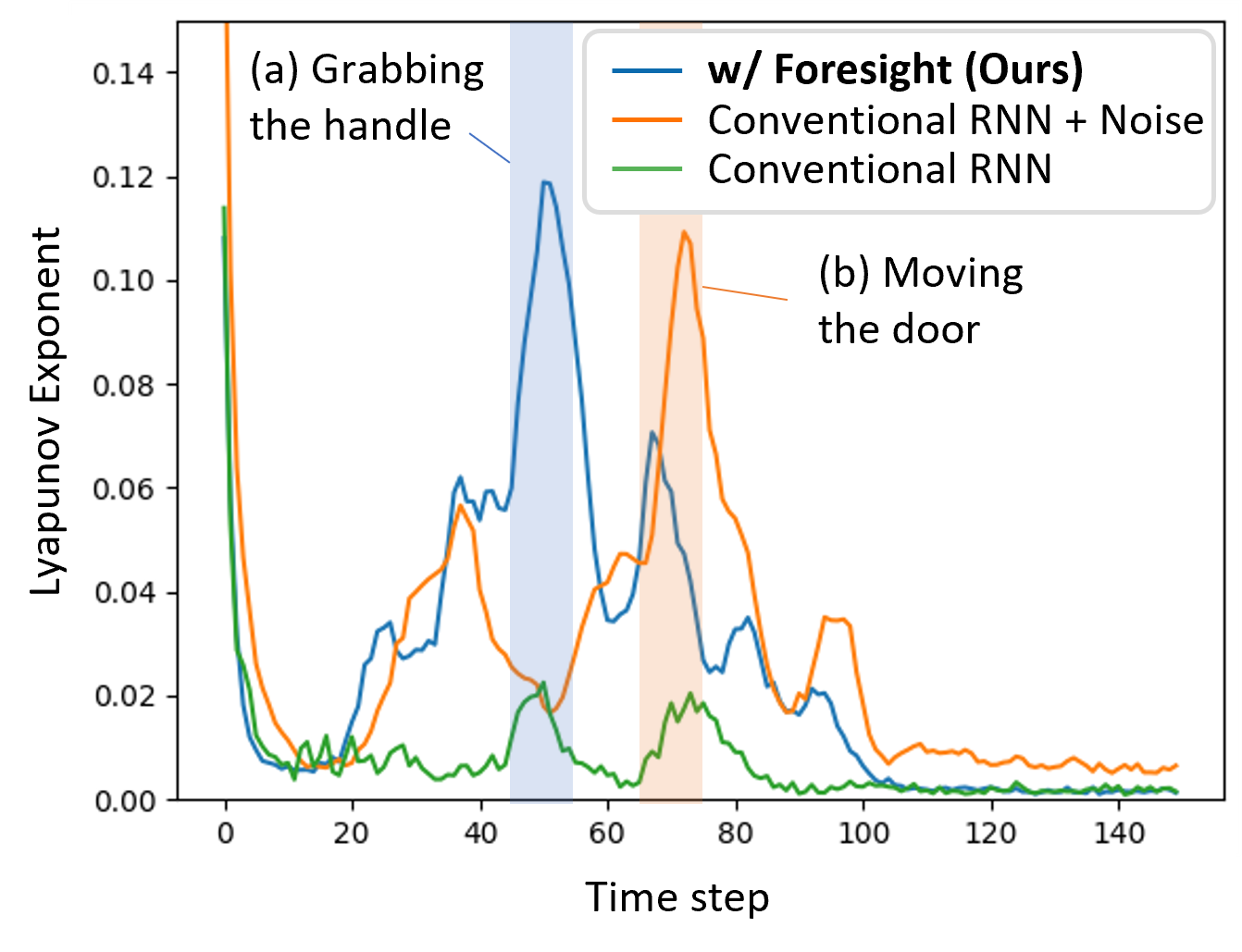}
    \caption{ Comparison of per-time-step Lyapunov exponents
    during motion prediction. Higher values represent that stronger
    chaotic properties are embedded at that timing.}
    \label{fig:lyapunov}
  \end{minipage}
\end{figure*}


The proposed model was evaluated on a door opening task as shown in Fig.~\ref{fig:env}.
The door can be opened one of three directions but cannot be visually distinguished.
The model was compared with a
conventional RNN model, and its variant which applies random noises to the hidden states
at each time step. The models were trained on five demonstrations of door opening motions
on each direction, using camera data and joint angle data as inputs.

The results showed that the proposed method was able to predict adaptive motions. The success
rates were over 80\% at the early stage of training, which is higher than those of fully trained
conventional models (see details in the video). The model was able to diverge its motion prediction
on all three motions, according to the presented door type. As Fig.~\ref{fig:pca} shows, the
trajectory of the RNN hidden states during online prediction fluctuated between those of different
motion types. This indicates that the proposed model was able to self-organize a hidden space
that can stably transition between different policy attractors based on the acquired feedback.

Conversely, conventional models failed in diverging between different motions, which only succeeded in
predicting two out of three motions at the most. This was likely caused by how
the model embedded the uncertainty in the RNN hidden states, as described in Fig.~\ref{fig:lyapunov}.
Fig.~\ref{fig:lyapunov} compares on Lyapunov exponents, which reflect the quantity of possible
divergence at each time step. The conventional RNN model had small values throughout the task,
indicating that small uncertainty, or chaotic properties, were embedded in the model. In contrast,
the proposed model and the noised variant showed clear peaks, at when the door handle was grabbed, and
when the door started to move, respectively. Such difference suggests that the proposed model embedded
the uncertainty on its policy, or the \textit{cause}, whereas the noised variant embedded on the
resultant observation, or the \textit{effect}. This is can be interpreted that the foresight prediction
biased the model to incorporate various future consequences that affect the policy selection. This
property is beneficial for implementing adaptive behaviors, because the construction of chaotic
attractors is essential for deriving diverse motions during exploration.



\bibliography{bibliography}

\begin{thebibliography}{1}

\bibitem{itosan}
Hiroshi Ito, Kenjiro Yamamoto, Hiroki Mori, and Tetsuya Ogata.
\newblock Efficient multitask learning with an embodied predictive model for door opening and entry with whole-body control.
\newblock {\em Science Robotics}, 7(65):eaax8177, April 2022.

\bibitem{srnn}
Shingo Murata, Jun Namikawa, Hiroaki Arie, Shigeki Sugano, and Jun Tani.
\newblock Learning to reproduce fluctuating time series by inferring their time-dependent stochastic properties: Application in robot learning via tutoring.
\newblock {\em IEEE Transactions on Autonomous Mental Development}, 5(4):298--310, 2013.

\bibitem{ai1}
Kai Ueltzhöffer.
\newblock Deep active inference.
\newblock {\em Biological Cybernetics}, 112:547--573, 2018.

\bibitem{pv_rnn}
Ahmadreza Ahmadi and Jun Tani.
\newblock {A Novel Predictive-Coding-Inspired Variational RNN Model for Online Prediction and Recognition}.
\newblock {\em Neural Computation}, 31(11):2025--2074, 11 2019.

\bibitem{diffusion_policy}
Cheng Chi, Siyuan Feng, Yilun Du, Zhenjia Xu, Eric Cousineau, Benjamin Burchfiel, and Shuran Song.
\newblock Diffusion policy: Visuomotor policy learning via action diffusion.
\newblock In {\em Proceedings of Robotics: Science and Systems (RSS)}, 2023.

\bibitem{predictive_learning}
Kanata Suzuki, Hiroshi Ito, Tatsuro Yamada, Kei Kase, and Tetsuya Ogata.
\newblock Deep predictive learning: Motion learning concept inspired by cognitive robotics, 2024.

\end{thebibliography}
\bibliographystyle{unsrt}

\end{document}